\documentclass[twoside,11pt]{article}
\pdfoutput=1
\usepackage{xcolor}

\usepackage{graphicx}
\graphicspath{ {./} }

\usepackage{dmlr2e}

\ShortHeadings{Towards Ontology-Enhanced Representation Learning for Large Language Models}{Francesco Ronzano and Jay Nanavati}

\firstpageno{1}

\usepackage{hyperref}

\begin{document}

\title{Towards Ontology-Enhanced Representation Learning for Large Language Models}

\author{\name Francesco Ronzano \email francesco.ronzano@iqvia.com \\
       \addr  IQVIA\\
       \AND
       \name Jay Nanavati \email jay.nanavati@iqvia.com \\
       \addr  IQVIA\\}

\maketitle

\begin{abstract}
Taking advantage of the widespread use of ontologies to organise and harmonize knowledge across several distinct domains, this paper proposes a novel approach to improve an embedding-Large Language Model (embedding-LLM) of interest by infusing the knowledge formalized by a reference ontology: ontological knowledge infusion aims at boosting the ability of the considered LLM to effectively model the knowledge domain described by the infused ontology. The linguistic information (i.e. concept synonyms and descriptions) and structural information (i.e. is-a relations) formalized by the ontology are utilized to compile a comprehensive set of concept definitions, with the assistance of a powerful generative LLM (i.e. GPT-3.5-turbo). These concept definitions are then employed to fine-tune the target embedding-LLM using a contrastive learning framework. To demonstrate and evaluate the proposed approach, we utilize the biomedical disease ontology MONDO. The results show that embedding-LLMs enhanced by ontological disease knowledge exhibit an improved capability to effectively evaluate the similarity of in-domain sentences from biomedical documents mentioning diseases, without compromising their out-of-domain performance.

\end{abstract}


\section{Introduction}
The availability of high-quality textual data is essential to boost the ability of LLMs to effectively understand, model and reason about the semantics of a text. With this respect, recently, \textit{Synthetic Textual Data Augmentation} methods have been gaining an increasing relevance: very large, pre-trained LLMs have been often exploited to generate, restructure or annotate textual data that in turns is exploited to enhance smaller, domain specific LLMs \citep{ding2024data, tan2024large}. Besides synthetic data, the structured information included in a diverse set of knowledge resources has also been used to boost LLMs giving rise to several proposals of \textit{Knowledge-resource Driven LLM-enhancement} techniques \citep{hu2023survey}: for example, knowledge graphs represent the most relevant type of knowledge resources exploited to this purpose \citep{yang2024give, yasunaga2022linkbert}.

Knowledge resources like ontologies are extensively used to organise and harmonize information inside and across a wide range of distinct domains and applications \citep{patel2024comprehensive}. Various methods have been proposed to improve machine learning model performance by relying on ontologies and vice-versa \citep{kulmanov2020machine}: recent examples include ontology-driven interaction with and fine-tuning of generative LLMs \citep{palagin2023ontochatgpt, baldazzi2023fine}, ontology-based refinement of knowledge graph queries \citep{allemang2024increasing} and exploitation of LLMs to create or enrich ontologies \citep{ciatto2024large, mateiu2023ontology}.

In this stream of works, the main contribution proposed and evaluated by this paper is \textbf{a novel, automated approach to infuse external knowledge, formalized by an ontology of interest, into an embedding-LLM (i.e. text encoder)}\footnote{Implementation available on GitHub repository: \url{https://github.com/iqvianlp/llm-onto-infuse/}.}. This is achieved by leveraging on both: (i) the linguistic and structural information formalized by an ontology (\textit{Knowledge-resource Driven LLM-enhancement}) and (ii) a powerful generative LLM (i.e. GPT-3.5-turbo) to perform \textit{Synthetic Textual Data Augmentation}. By using the generative LLM, a rich set of real and/or synthetic definitions are gathered for all the concepts specified by the considered ontology. These definitions are then exploited to create training samples useful to fine-tune a the target embedding-LLM by a contrastive learning framework: training samples (i.e. pairs of similar and dissimilar definitions) are generated by following a principled approach, aimed at maximizing their effectiveness for fine-tuning. Once fine-tuning finalizes, the vectorial representations of texts generated by the embedding-LLM will incorporate the knowledge formalized by the considered ontology. 

\section{Related work}
\label{rw}
Recently, several contrastive representation learning approaches have been proposed to improve the quality of text embeddings by exploiting collections of pairs of related or similar texts (e.g. query-answer, texts conveying the same meaning, etc.) to fine tune embedding-LLMs: the quality of embeddings is improved by increasing the similarity of the LLM-generated vectorial representations of semantically close texts \citep{hadsell2006dimensionality}. Examples of contrastive learning frameworks include Sentence-BERT \citep{reimers2019sentence} where a dual-encoder network architecture, coupled with multiple loss functions is used to fine-tune text embeddings in a supervised way. SimCSE \citep{gao2021simcse} proposes to use distinct LLM dropout masks as data augmentation strategies to generate pairs of similar emeddings for unsupervised fine-tuning of embedding-LLMs in a contrastive objective. \cite{schick2021generating} use generative LLMs to create labelled text pairs useful for unsupervised fine-tuning of embedding-LLMs. Also \cite{wang2023improving} relies on generative LLMs (i.e. GPT-3.5-turbo and GPT-4) to generate synthetic training data spanning over multiple tasks and languages: this data is then exploited to fine-tune Mistral-7b, a decoder-only LLM to generate better emeddings. \cite{su2022one} extend the text excerpt to be embedded with free-text instructions describing the task the embedding will be used for. Overall, a common paradigm exploited to fine-tune embedding-LLMs by contrastive learning relies on a contrastive pre-training phase that exploits collections of text pairs generated semi-automatically by weak supervision, followed by a fine-tuning phase where LLMs are improved by relying on higher-quality annotated datasets \citep{li2023towards, wang2022text}. Besides data augmentation approaches, two additional key ingredients useful to boost the effectiveness of contrastive learning frameworks exploited to generate better text embeddings are (i) the strategy to select of text pairs that will constitute training samples and (ii) the choice of the training objective (i.e. loss function) \citep{wang2023sncse}.
In \cite{liu2020self}, synonyms of biomedical concepts retrieved from the UMLS meta-thesaurus \citep{bodenreider2004unified} are exploited to fine-tune embedding-LLMs by contrastive learning. In comparison, our ontological knowledge infusion approach aims at enhancing embedding-LLMs by: (i) taking advantage of a richer set of linguistic and structural features of ontologies (beyond synonymy); (ii) exploiting the text of whole sentences (instead of noun phrases) to fine-tune embedding-LLMs; (iii) proposing a novel, automated approach to create training text pairs.

\section{Workflow to infuse ontological knowledge in embedding-LLMs}
\label{okiw}
The proposed approach to infuse ontological knowledge in embedding-LLMs relies on linguistic features - \textit{synonym terms} and \textit{definitions} - and structural features -  \textit{taxonomic relations} - that characterize the set of concepts defined by an ontology. As better detailed in the next Sections, these features, shared by most ontologies, are exploited to support and drive: (i) the \textit{generation of text excerpts describing the concepts of the ontology} and (ii) the \textit{effective aggregation of these text excerpts into pairs of similar or dissimilar ones}, to be exploited to fine-tune embedding-LLMs.

\subsection{Fine-tuning embedding-LLMs by contrastive learning}
\label{contr_learning_approach}
The the textual information (i.e. definitions of ontological concepts) gathered by relying on a reference ontology is infused in an embedding-LLMs of choice by fine-tuning such LLM in a contrsative objective.

\subsubsection{Contrastive learning architecture and training objective}
\label{training_obj}
In our experiments we rely on the contrastive learning framework described by \cite{chen2020simple} to perform ontological knowledge infusion. In principle, we can infuse ontological knowledge in any embedding-LLM ($EMB$) capable of generating, given a text $t$, the corresponding dense vectorial representation $h_{t} = EMB(t) \in R^n$, where $n$ represents the dimension of $h_{t}$, the text embedding\footnote{Depending on the specific scenario and embedding-LLM considered, the embedding $h_{t}$ can be generated by distinct strategies, including pooling of single-token embeddings or by considering special-purpose tokens of the embedding-LLM (e.g. the CLS token).}. Let suppose to have at our disposal a collection of $I$ pairs of semantically related texts $(t_{i}, t^+_{i})$, where $0 < i <= I$. Considering batches of $N$ pairs of semantically related texts, the corresponding pairs of embeddings $(h_{i}, h^+_{i})$ can be computed by relying on the LLM ($EMB$). The categorical cross-entropy loss is exploited to favour, for each embedding $h_{i}$, the identification of (i.e. prediction of the class associated to) the associated positive embedding $h^+_{i}$: samples from other embedding pairs in the same batch are considered as noise. This training objective, referred to as \textbf{InfoNCE loss} \citep{oord2018representation}, is described by the following formula:
\begin{equation}
\label{eq:1_infonce_loss}
loss_{i}=-log\frac{e^{sim(h_{i}, h^+_{i})/\tau}}{\sum_{i=1}^{N}e^{sim(h_{i}, h^+_{j})/\tau}}
\end{equation}
where $N$ is the batch size and $\tau$ is the temperature. $sim(p, q)$  is the similarity function between the embeddings $p$ and $q$: it is common practice to use cosine similarity.

It has been shown that when InfoNCE loss is exploited, the quality of learned embeddings improves sensibly if in each batch, for each pair of semantically related texts $(t_{i}, t^+_{i})$, one (or more) \textbf{hard negative texts} are included \citep{chen2017sampling, gao2021simcse}. Given a pair of positive texts, an associated hard negative sample $w$ is a text that is semantically distinct from the texts $t_{i}$ and $t^+_{i}$, even if its embedding $h_{w} = EMB(w)$ is characterized by a high semantic similarity with the embeddings of any positive text (i.e. $h_{i}$ and $h^+_{i}$). Therefore, if for each positive text pair one or more hard negative texts are selected, each training sample exploited by the considered contrastive learning framework and training objective would be represented by the tuple of texts $(t_{i}, t^+_{i}, w^{HN_{1}}_{i}, ..., w^{HN_{K}}_{i},)$ where $t_{i}$ and $t^+_{i}$ represent the pair of positive texts, while $w^{HN_{k}}_{i}$, with $0 < i <= K$, are the associated $K$ hard negative samples\footnote{If $K$ is zero, hard negative sampling is not exploited.}.

\subsubsection{Ontology-driven creation of training samples}
\label{training_data_creation}
This Section describes the novel procedure we devise to create training samples useful to infuse ontological knowledge into an embedding-LLM of our choice, in a contrastive objective. After generating synthetic definitions of the concepts included in the considered ontology by prompting a generative LLM, we exploit these definitions in order to create training samples, thus selecting positive text pairs as well as associated hard negative texts.

\begin{figure}[h]
\centering
\includegraphics[width=0.99\textwidth]{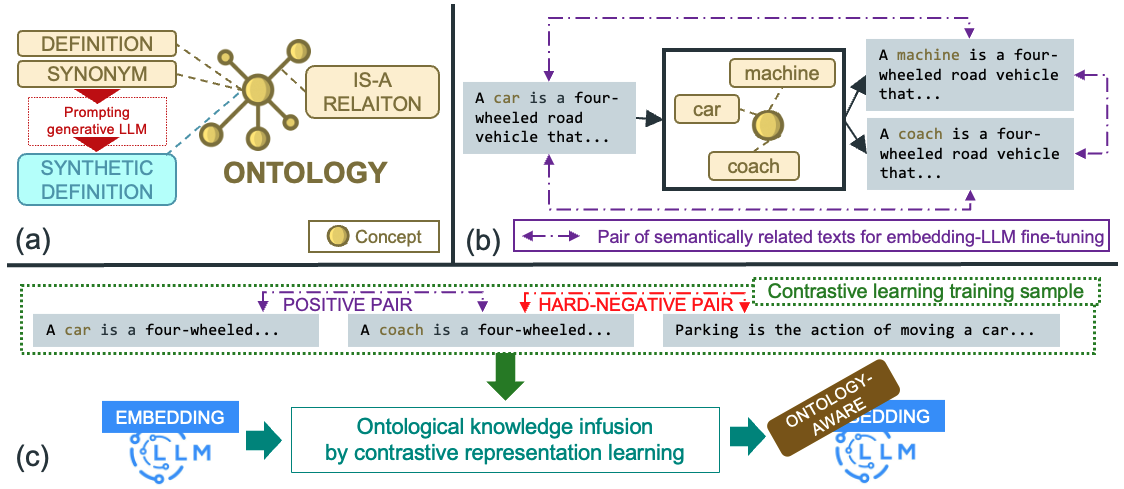}
\caption{(a) structure of ontologies, generation of synthetic concept definitions; (b) creation of pairs os semantically related sentences by synonym substitution; (c) overview of the ontological knowledge infusion approach.}
\label{fig:overall_procedure}
\end{figure}

\textbf{Prompting LLMs to generate synthetic concept definitions}: our ontological knowledge infusion approach is based on the availability of textual contents describing the concepts formalized by the ontology of choice: in the current setting, we focus on concept definitions\footnote{As future work, we would like to evaluate the knowledge infusion effectiveness of other types of textual information associated to concepts, distinct from definitions.}. Ontologies could include definitions of (part of) their concepts. To guarantee the availability of at least one definition associated to each concept, generative LLMs are prompted to create synthetic definitions: for each synonym of an ontology concept, a one-sentence synthetic definition of that concept is collected (see part (a) of Figure~\ref{fig:overall_procedure}). The structure of the definition-generation prompt is highly dependent on both the features and domain of the considered ontology and the generative LLM of choice.

\textbf{Creation of semantically related pairs of texts by synonym substitution}: as illustrated in Section~\ref{training_obj}, the contrastive learning framework exploited to infuse ontological knowledge in embedding-LLMs relies on training samples constituted by pairs of semantically related texts. To create such text pairs, we rely on both the definitions (real and synthetic ones) and the synonyms of the concepts of the ontology (see part (b) of Figure~\ref{fig:overall_procedure}).

For each concept, we select all its real or synthetic definitions that mention one and only one of its synonyms. Then for each selected definition, we generate similar definitions by replacing the mentioned synonym with a distinct synonym of the same concept. Therefore, in our contrastive learning settings, we consider as pairs of semantically related texts (i.e. training samples) all possible pairs of definitions of the same concept, one obtained by performing synonym substitution over the text of the paired definition\footnote{In our experiments, we observed that the exploitation of training samples that include pairs of definitions not obtained by synonym substitution - e.g. (real definition, synthetic definition) - hinders the performance of our ontological knowledge infusion approach, probably since both definitions would highlight distinct traits of the same concept, characterising the same concept through distinct perspectives.}. This approach would not allow the generation of any positive text pair for concepts with a single synonym: to remedy this, we automatically generate an additional, novel synthetic synonym for such concepts, by appending one of the synonyms of its parent concept to its actual synonym.

\textbf{Ontology-driven selection of hard negative samples}:
we rely on both embedding similarity and the taxonomic concept-to-concept relations specified by the ontology to select one hard negative text associated to each pair of semantically related texts (i.e. each pair of definitions of the same concept; see part (c) of Figure~\ref{fig:overall_procedure}). Given the positive text pair $(t_{i}, t^+_{i})$ including two definitions of the concept $C_{t}$, the associated hard negative text $w$ should verify the following conditions:

- considering the taxonomic relations defined by the ontology, the concept $C_{w}$ should not be an ancestor or a descendant of the concept $C_{t}$;

- the embedding of the definition $w$ (i.e. $h_{w}$) should represent, among all the embeddings of definitions of concepts distinct from $C_{t}$, the one with the highest semantic similarity with the embeddings of the texts $t_{i}$ and $t^+_{i}$ (i.e. $h_{t_{i}}$ and $h_{t^+_{i}}$)\footnote{Similarity scores are averaged across $t_{i}$ and $t^+_{i}$; we use cosine similarity in our experiments.}.

\section{Infusing disease knowledge relying on MONDO ontology}
\label{eval}
We showcase and evaluate our ontological knowledge infusion approach by infusing the disease knowledge formalized by a widespread and rich biomedical ontology into four distinct flavours of embedding-LLMs. More specifically, we consider MONDO\footnote{\url{https://mondo.monarchinitiative.org/}} \citep{vasilevsky2022mondo}, an ontology that aims at globally harmonizing the characterization of diseases by unifying the information contained in multiple knowledge resources and data models. We used the April-2024 version of MONDO that defines 24,201 disease-related concepts characterized by almost 75 thousands synonyms\footnote{The main name of each concept together with its EXACT synonyms are considered; obsolete concepts have been ignored.}. MONDO includes a definition of almost 70\% of its concepts. Concepts are hierarchically organized through 36,459 is-a relations.

We prompt GPT-3.5-turbo to generate a single-sentence definition from each synonym of concepts from the MONDO ontology (see Appendix~\ref{app:gpt_35_prompt}): 57,692 synthetic concept definitions are thus generated\footnote{This number is lower than the actual number of MONDO concept synonyms (about 75 thousands) since we applied specific synonym filtering rules resumed in Appendix~\ref{app:synonym_filtering_rules}.}, with at least one definition of each concept. 

By relying on the synonym substitution procedure described in Section~\ref{training_data_creation}, considering both ontology-provided and synthetic concept definitions, about 400 thousands pairs of semantically related definitions are generated. Each one of these pairs of definitions is extended with an hard negative definition selected by following the procedure explained in the latest part of Section~\ref{training_data_creation}: as a result, the ontology-driven creation of 400 thousand training samples ready to infuse the disease knowledge formalized by MONDO ontology in any embedding-LLM of choice is finalized.

We exploit the contrastive learning framework described in Section~\ref{training_data_creation} to infuse ontological knowledge in the following four embedding-LLMs:

- \textbf{PubMedBERT} \citep{gu2021domain}: a BERT-like LLM (110M parameters), pre-trained from scratch using abstracts from PubMed and full-text articles from PubMedCentral through masked token prediction and next sentence prediction;

- \textbf{SapBERT} \citep{liu2020self}: relies on PubMedBERT as base model, is fine-tuned in a contrastive learning framework to increase the similarity of pairs of synonyms of biomedical concepts, retrieved from the UMLS meta-thesaurus;

- \textbf{GTEbase} \cite{li2023towards}: an encoder LLM (110M parameters) created fine-tuning BERT-base-uncased by means of a two-stages contrastive learning framework: a pre-training phase relies on collections of text pairs generated semi-automatically by weak supervision. A subsequent training phase improves the LLM by higher-quality annotated datasets;

- \textbf{GIST} \citep{solatorio2024gistembed}: at the time of writing, represents one of the best performing small embedding-LLMs (about 100M) in the MTEB leader-board\footnote{\url{https://huggingface.co/spaces/mteb/leaderboard}}, fine-tuned by a contrastive objective relying on an dynamic selection strategy to identify in-batch negative samples.

Each one of these embedding-LLMs, all having a comparable number of parameters, is characterized by specific peculiarities that make it interesting with respect to our evaluations: \textbf{PubMedBERT} is not fine-tuned exploiting a contrastive objective and is pre-trained on texts from the same domain of the MONDO ontology; \textbf{SapBERT} is fine-tuned on biomedical-synonym knowledge encoded in the UMLS meta-thesaurus; \textbf{GTEbase} is fine-tuned on a huge set of annotated datasets, including biomedical ones; \textbf{GIST} is a robust LLM, highly-scoring in widespread embedding benchmarks. Appendix~\ref{app:fine_tuning_details} provides a detailed description of the LLM fine-tuning settings.

\subsection{Evaluation: datasets and results}
\label{eval_datasets}
We evaluate the quality of the embeddings generated by the four LLMs previously introduced, before and after infusing knowledge from the MONDO ontology. We choose sentence similarity (STS) to perform our evaluation since this task is commonly extensively used to measure the quality of text embeddings. We considered the following STS annotated datasets: \textbf{BIOSSES} \citep{souganciouglu2017biosses} including 100 sentence pairs from biomedical publications and \textbf{the five test sets of SemEval Sentence Similarity challenges} released yearly from 2012 to 2016 \citep{agirre2012semeval, agirre2013sem, agirre2014semeval, agirre2015semeval, agirre2016semeval}. Spearman's correlation is exploited to evaluate the performance of the four embedding-LLMs. Table~\ref{tab:boisses_results} and Table~\ref{tab:semeval_results} show BIOSSES  and SemEval results, respectively.
Appendix~\ref{app:datasets_info} provides detailed information on both datasets.

\begin{table}[ht]
\centering
\begin{tabular}{|p{3.3cm}||p{0.85cm}|p{0.85cm}|}
 \hline
 Embedding & \multicolumn{2}{|c|}{BIOSSES} \\
  LLM & All & Dis \\
 \hline
 $PubMedBERT_{orig}$ & 53.74 & 69.80 \\
 $PubMedBERT_{kinf}$ & \textbf{71.23} & \textbf{77.41} \\
 \hline
 $SapBERT_{orig}$    & 81.86 & 83.21 \\
 $SapBERT_{kinf}$    & \textbf{85.45} & \textbf{84.79} \\
 \hline
 $GTEbase_{orig}$    & 87.26 & \textbf{90.30} \\
 $GTEbase_{kinf}$    & \textbf{87.40} & 89.62 \\
 \hline
 $GIST_{orig}$       & 87.96 & 89.66 \\
 $GIST_{kinf}$       & \textbf{88.86} & \textbf{92.05} \\
 \hline
\end{tabular}
\caption{BIOSSES STS Spearman correlation scores, before ($orig$) and after ($kinf$) infusing disease-related ontological knowledge in embedding-LLMs. Evaluation scores computed considering: whole dataset (All), sentences with disease mentions (Dis).}
\label{tab:boisses_results}
\end{table}

\begin{table}[ht]
\centering
\begin{tabular}{|p{2.3cm}||p{0.85cm}|p{0.85cm}||p{0.85cm}|p{0.85cm}||p{0.84cm}|p{0.85cm}||p{0.85cm}|p{0.85cm}||p{0.85cm}|p{0.85cm}|}
 \hline
 Embedding & \multicolumn{2}{|c|}{STS12} & \multicolumn{2}{|c|}{STS13} & \multicolumn{2}{|c|}{STS14} & \multicolumn{2}{|c|}{STS15} & \multicolumn{2}{|c|}{STS16} \\
  LLM & All & Dis &  All & Dis &  All & Dis &  All & Dis &  All & Dis \\
 \hline
 $PMBERT_{orig}$  & 25.99 & 46.34 & 28.09 & 16.21 & 25.80 & 00.30 & 37.33 & 21.31 & 47.99 & \textbf{80.33} \\
 $PMBERT_{kinf}$  & \textbf{41.90} & \textbf{47.83} & \textbf{42.19} & \textbf{18.30} & \textbf{37.94} & \textbf{12.32} & \textbf{49.17} & \textbf{23.55} & \textbf{58.37} & 72.78 \\
 \hline
 $SapBERT_{orig}$     & 70.89 & 68.84 & 79.23 & 35.73 & 70.37 & 47.64 & 77.85 & 56.99 & 76.71 & 89.73 \\
 $SapBERT_{kinf}$     & \textbf{72.31} & \textbf{79.99} & \textbf{80.66} & \textbf{46.04} & \textbf{72.44} & \textbf{52.07} & \textbf{79.79} & \textbf{64.05} & \textbf{77.58} & \textbf{92.86} \\
 \hline
 $GTEbase_{orig}$     & 75.70 & 69.85 & 85.72 & 87.91 & 81.51 & 76.66 & 88.81 & 87.40 & 83.82 & 93.60 \\
 $GTEbase_{kinf}$     & \textbf{76.44} & \textbf{70.17} & \textbf{86.12} & \textbf{88.15} & \textbf{81.94} & \textbf{77.69} & \textbf{88.86} & \textbf{88.18} & \textbf{84.21} & \textbf{94.71} \\
 \hline
 $GIST_{orig}$        & 76.15 & 63.88 & 87.85 & 88.64 & 83.39 & 74.52 & 89.43 & \textbf{85.75} & 85.35 & \textbf{93.78}  \\
 $GIST_{kinf}$        & \textbf{76.69} & \textbf{65.94} & \textbf{87.99} & \textbf{89.26} & \textbf{83.64} & \textbf{75.45} & \textbf{89.56} & 85.42 & \textbf{85.69} & \textbf{93.78}  \\
 \hline
\end{tabular}
\caption{SemEval STS Spearman correlation scores, before ($orig$) and after ($kinf$) infusing disease-related ontological knowledge in embedding-LLMs. Evaluation scores computed considering: whole dataset (All), sentences with disease mentions (Dis). PMBERT is the abbreviated name for PubMedBERT.}
\label{tab:semeval_results}
\end{table}

\subsection{Discussion}
\label{eval_discussion}

Our ontological knowledge infusion approach consistently improves the sentence similarity performance of embedding-LLMs across a wide range of evaluation datasets, as demonstrated in Table~\ref{tab:boisses_results} and Table~\ref{tab:semeval_results}. By infusing disease knowledge, this novel method enhances both domain-specific (i.e. customized to effectively model biomedical texts) and general-purpose embedding-LLMs, with a more pronounced impact on domain-specific models like SapBERT. The infusion of domain knowledge improves \textit{in-domain} sentence similarity performance of embedding-LLMs, evaluated against biomedical sentence pairs (Table~\ref{tab:boisses_results} and 'Dis' columns of Table~\ref{tab:semeval_results}): at the same time, ontological knowledge infusion does not deteriorate the capability of embedding-LLMs to effectively evaluate the similarity of \textit{out-of-domain} sentences, from domains distinct from biomedicine ('All' columns of Table~\ref{tab:semeval_results}).

The four embedding-LLMs studied have comparable parameter counts, but the extent of improvement varies based on the pre-training and fine-tuning strategies they underwent before ontological knowledge infusion. LLMs built using more basic approaches, such as PubMedBERT, exhibit greater performance gains after ontological knowledge infusion. In contrast, more advanced LLMs like GTEbase and GIST show smaller but consistent improvements across most evaluation scenarios.
SapBERT, an embedding-LLM pre-trained on biomedical data using a novel, synonymy-based pre-training approach, demonstrates strong baseline performance. However, our ontological knowledge infusion method further enhances its sentence similarity capabilities, highlighting the effectiveness of our approach in improving even state-of-the-art domain-specific models.

\section{Conclusions and future work}
\label{concl}
In this paper we presented and evaluated a novel approach to infuse the knowledge formalized by ontologies in embedding-LLMs with the aim of improving the ability of such embedding-LLMs to effectively model the knowledge domain described by that ontology. We showcased the effectiveness of our approach by infusing the knowledge formalized by the disease ontology MONDO into four representative flavours of embedding-LLMs.

As future directions of research we would like to explore a wider range of scenarios and approaches useful to perform ontology-driven knowledge infusion in embedding-LLMs by: (i) comparing a wider range of LLM flavours, possibly with bigger sizes and distinct architectures; (ii) evaluating the effectiveness of ontological knowledge infusion when we consider ontologies describing distinct domains with different granularities; (iii) exploring alternative LLM-prompting strategies to generate textual data by relying on ontological knowledge; (iv) considering additional evaluation tasks, besides sentence similarity, thus making evaluations more comprehensive, spanning a wider set of usage scenarios of embedding-LLMs.

\section*{Reproducibility statement}
\label{reproducibility_statement}
The paper provides the information needed to reproduce the proposed ontological knowledge infusion approach and the related evaluations, in particular:
\begin{itemize}
    \item \textbf{Ontology-driven training sample creation procedure}: Section~\ref{training_data_creation} provides the procedural description of the distinct steps that contribute to the creation of training data to support ontological knowledge infusion. Related to this, Appendix~\ref{app:gpt_35_prompt} specifies the prompt used to generate synthetic definitions of ontological concepts by relying on GPT-3.5-turbo;
    \item \textbf{LLM fine-tuning information}: Section~\ref{training_obj} explains the fine-tuning approach we adopted, relying on a contrastive objective. Details concerning fine-tuning process and values of hyper-parameters are specified in Appendix~\ref{app:fine_tuning_details};
    \item \textbf{Evaluation datasets, metrics and procedure}: as described in Section~\ref{eval_datasets}, widespread public sentence similarity datasets (together with related standard evaluation metrics) are exploited to evaluate the effectiveness of ontology-driven knowledge infusion. More information on the contents of each evaluation dataset is provided in Appendix~\ref{app:datasets_info}.
\end{itemize}

To foster reproducibility of the results of this paper, the implementation of the proposed ontological knowledge infusion framework, together with the exploited evaluation procedures and references to evaluation datasets are available at \url{https://github.com/iqvianlp/llm-onto-infuse/}.

\impact{This work presents a novel approach to infuse knowledge from ontologies into embedding-based Large Language Models (LLMs), improving their ability to model and reason about the domain described by the ontology. By leveraging linguistic and structural information in ontologies, the proposed method generates concept definitions used to fine-tune LLMs through contrastive learning.
Potential positive impacts include enhancing LLMs' factual knowledge and reasoning capabilities in domains like biomedicine, law, and finance, supporting the development of reliable domain-specific applications. However, negative impacts may arise from amplifying biases and information unbalance, potentially in present in the ontologies. Care must be taken to use authoritative, regularly updated ontologies and to maintain human oversight, as LLMs can still make errors.
This promising approach could positively impact knowledge work across specialized domains, but responsible deployment practices are crucial. Further research can explore knowledge infusion from diverse ontologies into various LLM architectures to fully realize the potential benefits.}

\acks{The work presented in this paper has been supported by IQVIA.}

\vskip 0.2in
\bibliography{Ronzano__Towards_Ontology_Enhanced_Representation_Learning_for_Large_Language_Models}

\appendix
\section{GPT-3.5-turbo prompts to generate synthetic definitions}
\label{app:gpt_35_prompt}
In order to generate synthetic definitions of concepts from the MONDO ontology, for each synonym of each concept (i.e. for each \texttt{\textit{MONDO\_CONCEPT\_SYNONYM}}), GPT-3.5-turbo has been prompted by means of the following dialogue-prompt:
\begin{itemize}
    \item \texttt{\textbf{SYSTEM}: You are an expert in clinical and biomedical sciences.}
    \item \texttt{\textbf{USER}: Could you provide a single sentence with the definition of \textit{MONDO\_\newline CONCEPT\_SYNONYM}?}
\end{itemize}

\section{Fine-tuning hyper-parameters}
\label{app:fine_tuning_details}
Ontological knowledge infusion has been performed by fine-tuning the considered embedding-LLMs by relying on the Sentence Transformers Python module available at \url{https://sbert.net/}. Eembedding-LLMs have been fine-tuned by relying on the InfoNCE loss (temperature $\tau$ equal to 0.05) with the following hyper-parameters:
\begin{itemize}
    \item \textbf{batch size}: 24 training samples
    \item \textbf{learning rate}: 1e-8
    \item \textbf{learning rate scheduler}: constant learning rate after a warm-up period during which the learning rate increases linearly starting from 0 up to the set learning rate value (i.e. 1e-8)
    \item \textbf{weight decay}: 1e-4
\end{itemize}

Fine-tuning has been carried out for a maximum of two epochs: after the first epoch, the model with the best the Speareman correlation score computed against the whole BIOSSES dataset has been considered and fully evaluated.
Cosine similarity has been consider in order to quantify the distance between pairs of text embeddings.

\section{Detailed information on evaluation datasets}
\label{app:datasets_info}
Table~\ref{tab:sts_datasets_info} describes the size of the distinct Sentence Similarity evaluation datasets that have been exploited to evaluate the ontological knowledge infusion approach. For each dataset both the total number of sentence pairs and the number of sentence pairs where both sentences include one or more disease mentions is specified. Disease mentions have been spotted by relying on the disease Named Entity Recognition (NER) model introduced by \cite{ruas2022nilinker}, available at \url{https://huggingface.co/pruas/BENT-PubMedBERT-NER-Disease}.

\begin{table}[ht]
\centering
\begin{tabular}{|p{2cm}||p{4.5cm}|p{4.5cm}||}
 \hline
 Dataset & Total number of & Number of sentence pairs \\
  & sentence pairs (All) & mentioning diseases (Dis) \\
 \hline
 BIOSSES & 100 & 31  \\
 \hline
 STS12 & 3,108 & 34  \\
 STS13 & 1,500 & 17  \\
 STS14 & 3,750 & 59  \\
 STS15 & 3,000 & 36  \\
 STS16 & 1,186 & 15  \\
 \hline
\end{tabular}
\caption{Number of sentence pairs included in the BIOSSES and SemEval Sentence Similarity datasets: total number of pairs (All) and number of pairs where both sentences include one or more disease mentions (Dis).}
\label{tab:sts_datasets_info}
\end{table}

\section{Synonym filtering rules applied to MONDO ontology}
\label{app:synonym_filtering_rules}
We refine and filter the synonyms of concepts from MONDO ontology to be considered to generate synthetic concept definition by applying the following rules:
\begin{itemize}
    \item if any, text in parenthesis is deleted from all the synonyms;
    \item duplicate synonyms are removed (case-insensitively);
    \item for each concept, we consider just a single randomly-chosen synonym in case: (i) a pair of synonym is characterised by a string Levenshtein distance lower than 10 or (ii) a pair of synonyms differs just by word ordering (case- and punctuation-insensitively).
\end{itemize}

\end{document}